\documentclass[letterpaper, 10 pt, conference]{ieeeconf}  %

\IEEEoverridecommandlockouts                              %
\makeatletter
\let\NAT@parse\undefined
\makeatother

\usepackage{url}

\usepackage{makeidx}         %
\usepackage{graphicx}        %
\usepackage{multicol}        %
\usepackage{amsmath}
\usepackage{amssymb}
\usepackage{booktabs}
\usepackage{xspace}
\usepackage{color}
\usepackage[hang,flushmargin]{footmisc}
\usepackage{microtype}
\usepackage{float}
\usepackage{array}
\usepackage{multirow}
\usepackage{flushend}
\usepackage{svg}
\usepackage{makecell}
\usepackage{tabularx}
\usepackage{bm}
\usepackage{subcaption} %
\usepackage{stfloats} %

\usepackage{ifthen}
\newboolean{fourPageVersion}
\setboolean{fourPageVersion}{false}

\newboolean{anonymousSubmission}
\setboolean{anonymousSubmission}{false}

\ifbool{fourPageVersion}{
    \renewcommand{\baselinestretch}{0.99}
}{}

\usepackage{cite}
\usepackage{amsmath,amssymb,amsfonts}
\usepackage{algorithmic}
\usepackage{graphicx}
\usepackage{textcomp}
\usepackage{xcolor}
\usepackage{tensor}

\usepackage{booktabs}
\usepackage{outlines}
\usepackage{caption}
\usepackage{subcaption}
\usepackage[hang,flushmargin]{footmisc}
\usepackage{stmaryrd}

\usepackage{microtype}
\usepackage{hhline}
\usepackage{multirow}
\usepackage{siunitx}
\usepackage{makecell}
\usepackage{gensymb}
\usepackage{placeins}

\definecolor{cvprblue}{rgb}{0.21,0.49,0.74}
\usepackage[breaklinks,colorlinks]{hyperref}
\usepackage{cite}

\usepackage{xspace}
\newcommand{\eg}{\mbox{e.\,g.}\xspace}
\newcommand{\ie}{\mbox{i.\,e.}\xspace}

\renewcommand{\[}{\begin{equation}}
\renewcommand{\]}{\end{equation}}

\newcommand{\ourName}{AnchorD}
\newcommand{\ourNameFull}{Metric Grounding of Monocular Depth Using Factor Graphs}
\NewDocumentCommand{\ourNameTable}{o}{%
    \IfNoValueTF{#1}
    {\ourName{} (Ours)}
    {\ourName{}-{#1} (Ours)}
}
\newcommand{\datasetName}{SprayD}
\newcommand{\datasetNameTable}{\datasetName{} (Ours)}
\newcommand{\tableYes}{\cmark}
\newcommand{\tableNo}{\xmark}

\newcommand{\projectsite}{
    \ifbool{anonymousSubmission}{\texttt{anonymized-for-review}}{https://anchord.cs.uni-freiburg.de}%
}

\newcommand{\projectcode}{%
  \ifthenelse{\boolean{anonymousSubmission}}%
    {anonymized for review}%
    {projectcode.xyz}%
}

\newcommand{\R}{\ensuremath{\mathbb R}}
\newcommand{\Rgz}{\ensuremath{\R_{>0}}}

\newcommand{\height}{\ensuremath{H}}
\newcommand{\width}{\ensuremath{W}}
\newcommand{\widthResized}{\ensuremath{\width^\prime}}
\newcommand{\heightResized}{\ensuremath{\height^\prime}}

\newcommand{\numPatches}{\ensuremath{n}}
\newcommand{\patchSize}{\ensuremath{m}}

\newcommand{\rgb}{\bm I}

\newcommand{\depthnan}{\ensuremath{\varnothing}}

\newcommand{\depthSymb}{\ensuremath{D}}
\newcommand{\senText}{\text{sen}}
\newcommand{\mdeText}{\text{mde}}
\newcommand{\slopeText}{\text{slp}}

\NewDocumentCommand{\depth}{o}{
    \IfNoValueTF{#1}
    {\bm \depthSymb}
    {\depthSymb_{#1}}
}
\NewDocumentCommand{\depthSen}{o}{
    \IfNoValueTF{#1}
    {\bm \depthSymb^{\senText}}
    {\depthSymb^{\senText}_{#1}}
}
\NewDocumentCommand{\depthMde}{o}{
    \IfNoValueTF{#1}
    {\bm \depthSymb^{\mdeText}}
    {\depthSymb^{\mdeText}_{#1}}
}

\NewDocumentCommand{\depthRescaled}{o}{
    \IfNoValueTF{#1}
    {\bm \depthSymb^{\prime}}
    {\depthSymb^{\prime}_{#1}}
}
\NewDocumentCommand{\depthRescaledSen}{o}{
    \IfNoValueTF{#1}
    {\bm \depthSymb^{\prime \, \senText}}
    {\depthSymb^{\prime \, \senText}_{#1}}
}
\NewDocumentCommand{\depthRescaledMde}{o}{
    \IfNoValueTF{#1}
    {\bm \depthSymb^{\prime \, \mdeText}}
    {\depthSymb^{\prime \, \mdeText}_{#1}}
}

\NewDocumentCommand{\depthOptimized}{o}{
    \IfNoValueTF{#1}
    {\bm \depthSymb^{\star}}
    {\depthSymb^{\star}_{#1}}
}

\NewDocumentCommand{\depthRescaledPredicted}{o}{
    \IfNoValueTF{#1}
    {\bm{\hat{\depthSymb^\prime}}}
    {\hat{\depthSymb^\prime}_{#1}}
}

\NewDocumentCommand{\depthPredicted}{o}{
    \IfNoValueTF{#1}
    {\bm{\hat{\depthSymb}}}
    {\hat{\depthSymb}_{#1}}
}

\newcommand{\slopeSymb}{\ensuremath{s}}
\NewDocumentCommand{\slope}{o}{
    \IfNoValueTF{#1}
    {\bm \slopeSymb}
    {\slopeSymb_{#1}}
}

\NewDocumentCommand{\slopeOptimized}{o}{
    \IfNoValueTF{#1}
    {\bm \slopeSymb^{\star}}
    {\slopeSymb^{\star}_{#1}}
}

\newcommand{\biasSymb}{\ensuremath{b}}
\NewDocumentCommand{\bias}{o}{
    \IfNoValueTF{#1}
    {\bm \biasSymb}
    {\biasSymb_{#1}}
}

\NewDocumentCommand{\biasOptimized}{o}{
    \IfNoValueTF{#1}
    {\bm \biasSymb^{\star}}
    {\biasSymb^{\star}_{#1}}
}

\newcommand{\mdeModel}{\ensuremath{f^{\mdeText}}}

\newcommand{\pixelDomainSymb}{\ensuremath{\Omega}}
\newcommand{\pixelDomainResized}{\ensuremath{\bm \pixelDomainSymb^\prime}}

\newcommand{\mdeFactor}{\phi^{\mdeText}}
\newcommand{\senFactor}{\phi^{\senText}}
\newcommand{\slopeFactor}{\ensuremath{\phi^{\slopeText}}}

\newcommand{\weightSymb}{\ensuremath{\lambda}}
\newcommand{\weightSen}{\ensuremath{\weightSymb^{\senText}}}
\newcommand{\weightMde}{\ensuremath{\weightSymb^{\mdeText}}}
\newcommand{\weightSlope}{\ensuremath{\weightSymb^{\slopeText}}}

\newcommand{\costSymb}{\ensuremath{C}}
\newcommand{\cost}{\ensuremath{\costSymb}}

\newcommand{\indicator}[1]{\ensuremath{\mathbb{I} [#1]}}
\newcommand{\huber}{\ensuremath{\rho}}
\newcommand{\huberThreshold}{\ensuremath{\delta}}
\newcommand{\huberThresholdMdeSen}{\ensuremath{\huberThreshold_1}}
\newcommand{\huberThresholdSlope}{\ensuremath{\huberThreshold_2}}

\newcommand{\neighbourSet}{\ensuremath{\mathcal{N}}}

\newcommand{\numSamplesInit}{\ensuremath{k}}

\newcommand{\patchCenterSymb}{\ensuremath{c}}
\newcommand{\patchCenter}[1]{
    \ensuremath{\patchCenterSymb_{#1}}
}

\newcommand{\gaussSymb}{\ensuremath{g}}
\newcommand{\gauss}[1]{
    \ensuremath{\gaussSymb\left(#1\right)}
}

\usepackage[capitalize]{cleveref}
\crefname{section}{Sec.}{Secs.}
\Crefname{section}{Section}{Sections}
\Crefname{table}{Table}{Tables}
\crefname{table}{Tab.}{Tabs.}
\crefname{algorithm}{Algo.}{Algos.}
\Crefname{algorithm}{Algorithm}{Algorithms}

\crefname{section}{Sec.}{Secs.}
\Crefname{section}{Section}{Sections}

\crefname{figure}{Fig.}{Figs.}
\Crefname{figure}{Figure}{Figures}

\crefname{table}{Tab.}{Tabs.}
\Crefname{table}{Table}{Tables}

\crefname{algorithm}{Algo.}{Algos.}
\Crefname{algorithm}{Algorithm}{Algorithms}

\crefname{subappendix}{App.}{Apps.}
\Crefname{subappendix}{Appendix}{Appendices}

\definecolor{red}{RGB}{255, 0, 0}   %
\definecolor{orange}{RGB}{255, 77, 0}   %
\definecolor{green}{RGB}{0, 128, 0}   %
\definecolor{purple}{RGB}{160, 32, 240}   %
\definecolor{lightblue}{RGB}{52, 155, 235}   %
\definecolor{darkmagenta}{RGB}{204, 51, 139} %

\def\BibTeX{{\rm B\kern-.05em{\sc i\kern-.025em b}\kern-.08em
    T\kern-.1667em\lower.7ex\hbox{E}\kern-.125emX}}

\usepackage{pifont}%

\hypersetup{
    colorlinks=true,
    linkcolor=blue,
    filecolor=magenta,    
    citecolor=green,
    urlcolor=blue,
    pdftitle={\ourName: \ourNameFull},
}

\usepackage[export]{adjustbox}

\usepackage{pifont}
\newcommand{\cmark}{\ding{51}} %
\newcommand{\xmark}{\ding{55}} %

\title{\LARGE \bf
    \ourName: \ourNameFull
}

\ifbool{anonymousSubmission}{
    \author{Anonymous Author(s)}
}{
    \author{Simon Dorer$^{1,2}$, Martin Büchner$^{1}$, Nick Heppert$^{1,2}$, Abhinav Valada$^{1}$
    \thanks{$^1$Department of Computer Science, University of Freiburg, Germany.}%
    \thanks{$^2$Zuse School ELIZA}%
    \thanks{This work was partially funded by the Carl Zeiss Foundation with the ReScaLe project. Dorer and Heppert are supported by the Konrad Zuse School of Excellence in Learning and Intelligent Systems (ELIZA) through the DAAD programme Konrad Zuse Schools of Excellence in Artificial Intelligence, sponsored by the Federal Ministry of Education and Research.}
    }
}

\begin{document}

\maketitle

\begin{abstract}
   Dense and accurate depth estimation is essential for
robotic manipulation, grasping, and navigation, yet currently available depth sensors are prone to errors on transparent, specular, and general non-Lambertian surfaces. 
To mitigate these errors, large-scale monocular depth estimation approaches provide strong structural priors, but their predictions can be potentially skewed or mis-scaled in metric units, limiting their direct use in robotics.
Thus, in this work, we propose a training-free depth grounding framework that anchors monocular depth estimation priors from a depth foundation model in raw sensor depth through factor graph optimization. Our method performs a patch-wise affine alignment,  locally grounding monocular predictions in metric real-world depth while preserving fine-grained geometric structure and discontinuities. To facilitate evaluation in challenging real-world conditions, we introduce a benchmark dataset with dense scene-wide ground truth depth in the presence of non-Lambertian objects. Ground truth is obtained via matte reflection spray and multi-camera fusion, overcoming the reliance on object-only CAD-based annotations used in prior datasets. Extensive evaluations across diverse sensors and domains demonstrate consistent improvements in depth performance without any (re-)training. We make our implementation publicly available at \url{https://anchord.cs.uni-freiburg.de}.

\end{abstract}

\section{Introduction}
\label{sec:intro}

Dense metric depth is a core requirement for robotic perception and interaction, including grasping \cite{cleargrasp,depthGrasp,transcg,sundermeyer2021contact}, manipulation~\cite{heppert2024ditto, von2024art}, localization~\cite{cattaneo2020cmrnetpp}, and scene understanding~\cite{bevsic2022dynamic, buechner2026momasg, valada2016towards}. However, acquiring reliable depth remains challenging in the presence of transparent, specular, and other non-Lambertian surfaces~\cite{transcg, werby2025arti}. In such regions, refraction, specular reflection, multipath effects, and violations of photometric correspondence assumptions often cause common depth sensors to yield invalid, sparse, or severely corrupted measurements. As a result, robots operating in real environments frequently lack accurate and precise geometry in regions that are most critical for safe manipulation and navigation~\cite{buechner2026momasg}.\looseness=-1

To counter this problem, several methods employ global metric scaling of monocular depth estimates to obtain dense depth maps~\cite{patel2025robotic,li2025novaflow, chen2025vidbot}. However, this often yields depth errors in the cm-range due to skew and bias in monocular depth estimates. Similarly, metric depth foundation models still lack the required accuracy to estimate precise explicit models from scene geometry, \eg, to identify grasps, or execute manipulation policies that require depth inputs. Although real-world depth sensors often fail on non-Lambertian surfaces, the corresponding RGB image typically retains informative cues, such as edges, textures, and semantic structure. Therefore, RGB-guided depth grounding provides a natural framework for recovering dense depth from sparse, missing, or corrupted sensor measurements. Existing methods, however, address this problem through supervised learning~\cite{lifd,transcg,todd,zhai2025tcrnet}. While effective under matched training and test conditions, such approaches require ground truth depth annotations that are particularly difficult to obtain for transparent and reflective objects in real scenes. Moreover, learned models are often sensitive to domain shift across sensors, materials, environments, and capture setups, making adaptation to new scenarios dependent on retraining or fine-tuning. In robotic deployment, where task-specific annotated data may be unavailable, this requirement can substantially limit practical applicability.\looseness=-1

\begin{figure}
    \vspace{0.2cm}
    \centering
    \includegraphics[width=\linewidth]{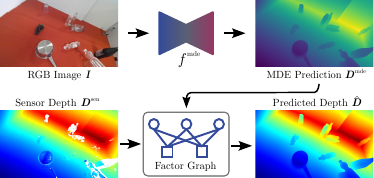}
    \caption{\textbf{Approach overview.} We present \ourName{} as a training-free method for grounding monocular depth predictions (MDE) using real-world sensor depth using a factor graph approach, which yields grounded, highly-accurate depth predictions on typical non-Lambertian objects.}
    \label{fig:approach-overview}
    \vspace{-0.5cm}
\end{figure}

In this work, we propose a zero-shot depth grounding framework that avoids task-specific training altogether. Our method combines raw metric sensor depth with monocular depth estimation (MDE) priors from off-the-shelf foundation models and formulates their fusion as a factor-graph optimization problem. By introducing patch-wise affine alignment, our method locally grounds monocular depth predictions in metric space while preserving structural cues from the monocular prior. This yields a flexible, training-free, and zero-shot approach applicable across different sensors and domains without requiring camera intrinsics or additional supervision. An overview of our approach is provided in \cref{fig:approach-overview}.\looseness=-1

A second challenge is the limited availability of suitable benchmarks for this problem. Existing datasets for depth completion of transparent or non-Lambertian objects are typically either synthetic, in which realistic sensor noise patterns are difficult to reproduce faithfully, or real-world datasets that provide ground truth depth only within the object regions. Such formulations are insufficient for robotics, where accurate geometry of the surrounding scene is also required, including supporting surfaces and nearby obstacles. To address this gap, we introduce a novel real-world dataset and data collection pipeline that provides dense ground truth depth maps across nearly the entire image. Our setup relies on an industrial-grade matte diffuser spray to produce stable reflections and performs multi-camera fusion, yielding scene-wide ground truth and enabling evaluation in the presence of transparent and non-Lambertian objects.

To summarize, our main contributions are as follows:
\begin{outline}
    \1 A novel factor graph-based framework for depth completion that grounds MDE foundation models.
    \1 A dataset for non-Lambertian object depth completion with dense ground truth depth, even in areas outside of the objects of interest.
    \1 GPU-parallelized implementation of the framework using jaxLS 
    and the full dataset, available at \projectsite.
\end{outline}

\section{Related Work}
\label{sec:related_work}
\label{subsec:related_work:depth_completion}

\noindent{\textbf{Depth Completion:}} Image-guided depth completion methods can broadly be grouped into two categories: approaches that operate on sparse but largely reliable sensor depth maps~\cite{sparse-completionformer,sparse-CostDCNet,sparse-cspn,sparse-dyspn,sparse-nlspn,sparse-sd2}, and approaches designed for transparent objects~\cite{cleargrasp,lifd,transcg,todd,tod,zhai2025tcrnet}, where depth measurements are not only incomplete but often heavily corrupted within object regions.
Among the former, DepthPrompting~\cite{depthPrompting} reconstructs dense depth from sparse measurements by leveraging a foundation monocular depth model together with a learned prompt-and-propagation module for sensor-agnostic sparse depth completion. In contrast, our method formulates the fusion of monocular priors and raw sensor observations as a training-free factor-graph optimization problem. It is specifically designed for scenarios in which sensor depth is not merely sparse, but can also be severely corrupted, as is commonly the case for transparent and other non-Lambertian surfaces.

Transparent-object-focused approaches explicitly address the characteristic failure modes caused by refraction and reflection. Early approaches such as ClearGrasp~\cite{cleargrasp} and DepthGrasp~\cite{depthGrasp} therefore explicitly detect transparent regions, remove unreliable depth measurements, and recover geometry from RGB cues such as surface normals and occlusion boundaries. More recent methods move toward direct RGB-D prediction: LIFD~\cite{lifd} models local implicit depth functions for transparent-object completion, TranspareNet~\cite{todd} refines transparent-region geometry in both 3D point clouds and image space. DFNet~\cite{transcg} and TCRNet~\cite{zhai2025tcrnet} learn end-to-end depth completion networks without explicitly incorporating geometric cues. In contrast to these supervised and task-specific approaches, our method does not learn a transparent-object-specific completion model, but instead combines a pretrained monocular depth prior with noisy sensor observations in an optimization framework.

\noindent{\textbf{Monocular Depth Estimation (MDE):}} Recent advances in large-scale training and modern model architectures have substantially improved monocular depth estimation (MDE), yielding models with strong zero-shot generalization across diverse scenes~\cite{dav2,dav3,metric3Dv2,depthPro,unidepth,unidepthv2}.  A central distinction in recent MDE is between methods that predict depth only up to an unknown global transformation and methods that directly infer metric depth from a single RGB image. Models such as DepthAnythingV2~\cite{dav2} and DepthAnything3-Mono~\cite{dav3} provide strong dense geometric priors, but their predictions remain ambiguous up to a global affine transformation. In parallel, recent methods have increasingly targeted metric monocular depth estimation directly~\cite{metric3Dv2,depthPro,unidepth,unidepthv2,dav3}. In this work, we follow the former direction and use DepthAnything3-Mono~\cite{dav3} as a dense prior, which we subsequently anchor in metric space using noisy sensor depth within our optimization framework.\looseness=-1

\noindent{\textbf{Depth Completion Datasets:}} Classical depth completion is commonly evaluated on NYUv2~\cite{nyuv2} and KITTI~\cite{kitti}, which serve as standard indoor and outdoor benchmarks, respectively. For transparent-object depth completion, dedicated datasets are required, since commodity depth sensors typically fail precisely on transparent surfaces. Existing datasets can be broadly categorized into physical and virtual approaches. Physical approaches obtain supervision by replacing transparent objects with opaque counterparts, as done in TOD~\cite{tod} and ClearGrasp~\cite{cleargrasp}. Virtual approaches instead rely on CAD models in combination with pose estimation, tracking, or geometric alignment, as in TODD~\cite{todd}, TransCG~\cite{transcg}, and ClearPose~\cite{clearpose}. Despite these differences in acquisition, both categories usually provide annotations only for regions containing transparent or other non-Lambertian objects, rather than dense scene-level ground truth corresponding to the physically nearest surface in cluttered real-world environments.

\section{\datasetName{} Dataset}
\label{sec:dataset}

In this section, we introduce \textit{\datasetName{}}, a novel and dense dataset for depth completion in scenes containing non-Lambertian and transparent objects. The key to our data collection is an industry-grade matte spray that enables accurate observations with off-the-shelf depth cameras.

\noindent{\textbf{Comparison to Existing Datasets:}} Compared to existing datasets, \datasetName{} provides two key contributions. First, while prior datasets typically restrict ground truth depth supervision to object masks~\cite{clearpose,transcg}, \datasetName{} provides dense depth annotations across the entire scene, including both non-Lambertian object regions and the surrounding background. This design choice is critical for downstream robotic perception tasks, which require accurate geometric understanding not only of target objects but also of supporting structures such as tables, shelves, and walls that constrain navigation and manipulation.

Second, in existing datasets such as ClearPose~\cite{clearpose}, objects occluded by transparent or translucent covers are often annotated with the depth of the underlying object rather than that of the physically closest surface, i.e., the cover itself. In contrast, we frame the problem from a real-world robotic perspective, in which robots must reason about physical workspace accessibility and potential obstructions during interaction. Accordingly, we define depth at such locations as the depth of the closest physical surface along the camera ray, rather than the depth of the spatially occluded but still visible object. 
In contrast to existing datasets~\cite{clearpose}, our proposed dataset does not require digital twins or CAD models for dense depth annotations. This also means that the proposed dataset collection procedure is readily extendable to larger objects, such as windows, glass tables, and various cluttered objects. We compare existing datasets in \cref{tab:dataset-comparison}.

\begin{table}
    \centering
    \resizebox{\columnwidth}{!}{%
    \begin{tabular}{l|cccc}
        \toprule
        Dataset & \# Frames & \# Depth sensing types & No digital/physical twin \\
        \midrule
        \datasetNameTable            & 398  & 3 & \tableYes \\
        Clearpose \cite{clearpose}   & 350K & 1 & \tableNo \\
        ClearGrasp \cite{cleargrasp} & 286  & 1 & \tableNo  \\
        TOD \cite{tod}               & 48K  & 1 & \tableNo \\
        TODD \cite{todd}             & 15K  & 1 & \tableNo \\
        TransCG \cite{transcg}       & 58k  & 2 & \tableNo  \\
        \bottomrule
    \end{tabular}}
    \caption{\textbf{Overview of various depth completion datasets.} While \datasetName{} is considerably smaller than previous datasets it yields true real-world ground truth depth maps and provides three unique real-world depth cameras and sensing types, being the Intel RealSense (Infra-Red Stereo), the ZED2 (RGB Stereo), and the Azure Kinect DK (Infra-red Time-of-Flight).}
    \label{tab:dataset-comparison}
    \vspace{-0.4cm}
\end{table}

\begin{figure}
        \centering
        \includesvg[width=0.7\linewidth]{assets/figures/dataset_cam_pointcloud_rgb.svg}
        \caption{\textbf{Dataset camera setup.} We employ three cameras: two RGB-D sensors, an Azure Kinect DK (green), and an Intel RealSense D415 (yellow), as well as a ZED2 stereo camera (blue). The point clouds captured by all cameras are registered in a shared reference frame and fused to generate ground truth depth.}
        \label{fig:dataset-cam-pointcloud-rgb}
        \vspace{0.5cm}
        \includegraphics[width=0.7\linewidth]{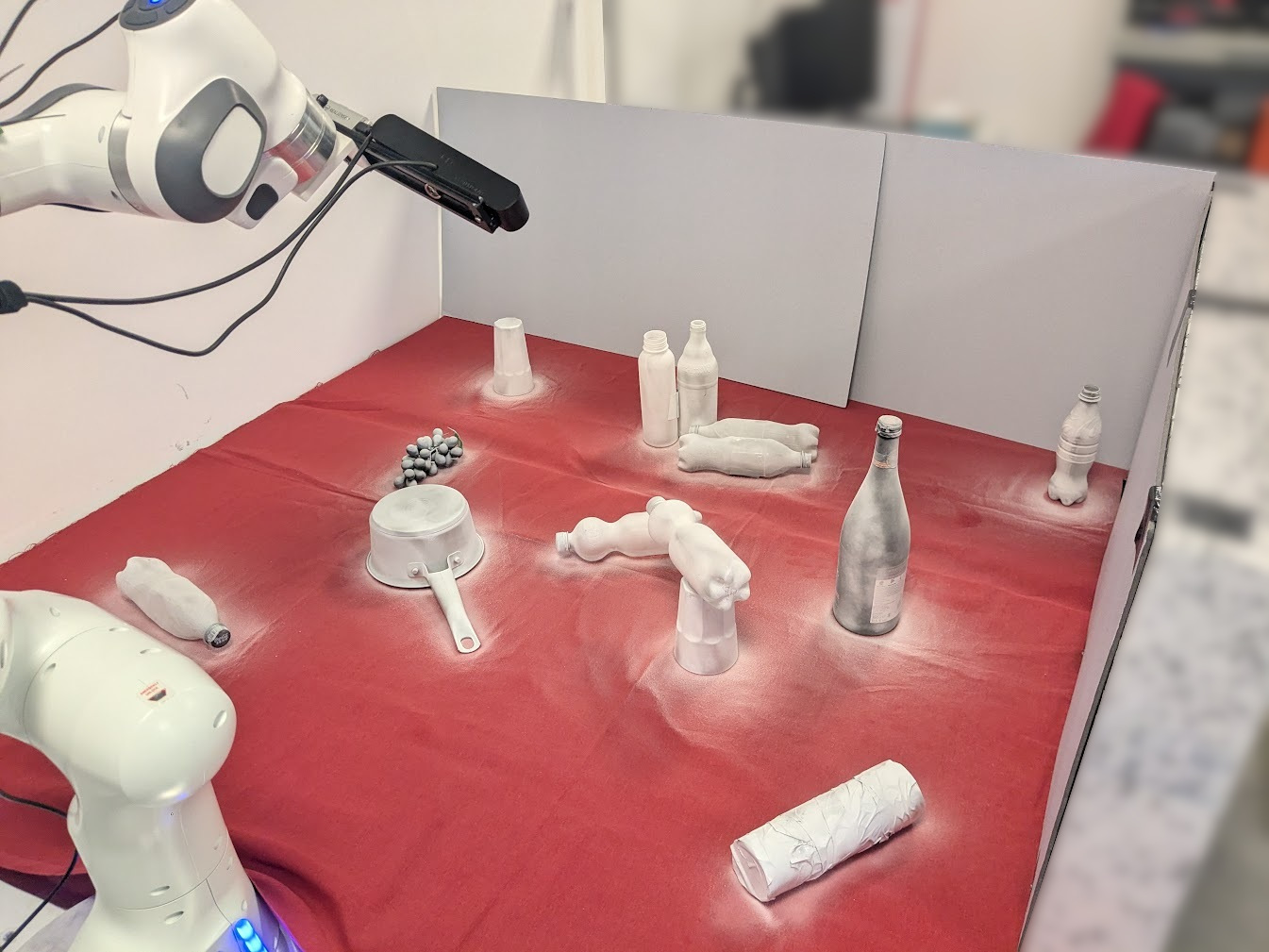}
        \caption{\textbf{Dataset collection.} We visualize one of the recorded scenes with the industrial-grade matte diffuser spray applied.}
        \label{fig:collection_spray}
    \vspace{-0.4cm}
    \end{figure}

\noindent{\textbf{Data Collection:}}
For data collection, we use three camera systems that provide synchronized color and depth: an Intel RealSense D415, a Stereolabs ZED2, and a Microsoft Azure Kinect DK RGB-depth camera, each operating at a native resolution of \(720 \times 1280\) pixels. By combining these three sensors, the dataset intentionally captures different depth-sensing modalities, depth accuracy regimes, failure modes, and field of views. 
As depicted in \cref{fig:dataset-cam-pointcloud-rgb}, all three cameras are rigidly mounted on a single camera rig attached to a robot arm. Using a robot arm provides two key advantages for data collection: (a) the process can be easily scaled through partial automation. (b) more importantly, it enables executing identical camera pose trajectories across multiple recording passes allowing us to capture the same scene geometry twice: once to record raw sensor data and once to acquire data suitable for ground truth generation.\looseness=-1

For each scene in the dataset, we placed a varying set of non-Lambertian objects on a table and followed a predefined trajectory with fixed stopping poses. At each pose, we captured synchronized RGB and depth images across all three cameras. These recordings constitute the raw samples of the dataset. Subsequently, we applied a matte diffuse spray to all non-Lambertian surfaces in the scene as depicted in \cref{fig:collection_spray} and repeated the same camera trajectory to obtain aligned RGB-depth images of the sprayed scene.\looseness=-1

\begin{figure*}
    \centering
    \begin{minipage}[c]{0.9\linewidth}
        \centering
        \includesvg[width=\linewidth]{assets/figures/depth_evolution.svg}
    \end{minipage}
    \hfill
    \begin{minipage}[c]{0.07\linewidth}
        \centering
        \includegraphics[width=\linewidth]{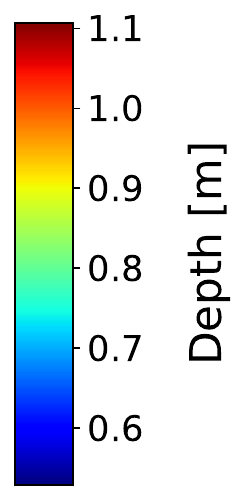}
    \end{minipage}
    \caption{\textbf{Ground truth depth generation process.} Depth obtained from the raw scene (left), after applying a diffuse spray to non-Lambertian objects in the scene (middle), and after fusing the depth maps from all three cameras in a common reference frame (right).}
    \label{fig:dataset-depth-evolution}
    \vspace{-0.5cm}
\end{figure*}

\noindent{\textbf{Ground Truth Post-Processing:}}
The depth maps captured from the sprayed scenes are generally accurate, but occasional miss values near object boundaries and in occluded regions, due to depth-to-RGB alignment. To mitigate these artifacts, we fuse depth information across all cameras into each individual camera frame. 
Using factory-provided intrinsics and calibrated extrinsics, we project each depth image from a camera into the image frame of all other cameras. When multiple depth candidates map to the same pixel, we retain the depth value associated with the point closest to the camera center, thereby reflecting our goal of \textit{workspace depth}. As a final post-processing step, we remove depth values exceeding the scene bounds (\( 1.7m \)) and manually filter misaligned or spurious samples. 
Following this procedure, we achieve 97.8\% dense ground truth maps, whereas ClearPose~\cite{clearpose} is at 19.4\%.
For fine-grained evaluation, we additionally generate object masks using Grounding DINO~\cite{grounded-dino} and SAM2~\cite{sam,sam2}, using the text prompt \verb|foreground objects| to obtain class-agnostic object segments.%

\section{Technical Approach}
\label{sec:technical-approach}
In this section, we introduce \ourName, a novel factor-graph-based optimization method for zero-shot monocular depth completion in scenes containing non-Lambertian surfaces.
Formally, given an RGB image \( \rgb \in \R^{\height \times \width \times 3} \) and an aligned corresponding raw sensor metric depth map \( \depthSen \in (\Rgz \cup \{\depthnan\})^{\height \times \width} \), where missing or invalid measurements are indicated by \( \depthnan \)%
\footnote{It is not necessary that the depth map is produced by the same sensor as the RGB image, as long as the depth map can be aligned with the RGB image through extrinsic and intrinsic calibration.}.
Our goal is to infer a dense, metrically accurate depth map \( \depthPredicted \in \Rgz^{\height \times \width} \).
At its core, \ourName{} formulates depth completion as a factor graph estimation problem that jointly optimizes dense depth values and patch-wise affine alignment parameters between monocular depth predictions and metric sensor measurements, enabling a locally-adaptive fusion of both sources.
An overview of our approach is shown in \cref{fig:approach-overview}.

\subsection{MDE Prediction}
\label{sec:mde-prediction}
At first, we compute a dense monocular depth estimate from the input image \( \rgb \) using a monocular depth estimation (MDE) foundation model \( \mdeModel \), e.g. DepthAnything3~\cite{dav3}:
\[
\depthMde = \mdeModel(\rgb) \in \Rgz^{\height \times \width}
\]
As it is common in monocular depth estimation, MDE models only provide depth estimates up to an unknown affine transformation, \ie, they are unscaled.

\subsection{Patch-wise Alignment}
\label{patch-wise-alignment}
To allow for a spatially varying alignment between monocular and optimized depth, we partition \( \depthMde \) into a regular grid of $\numPatches$ non-overlapping square patches of size \( \patchSize \times \patchSize \). Each patch is later associated with an independent set of affine alignment parameters, enabling local adaptation to depth distortions within the monocular depth prediction. 
In contrast to learned superpixels, fixed-size patches remain robust under extremely sparse or noisy sensor observations. This is particularly important in non-Lambertian regions, where depth measurements are unreliable and can cause adaptive segmentations to overfit to spurious sensor artifacts. Using regularly spaced patches, therefore, provides a stable spatial support for local affine alignment while maintaining computational simplicity.

To ensure that the \( \depthMde \) can be evenly decomposed into patches, we resize both the sensor depth map \( \depthSen \) and the monocular depth prediction \( \depthMde \) using nearest-neighbor interpolation such that their spatial dimensions are integer multiples of \( \patchSize \):
\[
\depthRescaledSen, \depthRescaledMde \in \mathbb{R}_{>0}^{H^\prime \times W^\prime},
\]
where
\(
\heightResized = \left\lfloor \tfrac{\height}{\patchSize} \right\rfloor \patchSize,
\)
and \(
\widthResized = \left\lfloor \tfrac{\width}{\patchSize} \right\rfloor \patchSize.
\)

We define the discrete image domain of the resized depth maps as
\[
\pixelDomainResized = \{1, \ldots, \heightResized\} \times \{1, \ldots, \widthResized\}.
\]
and denote the domain of the \(i\)-th superpixel by
\(
\pixelDomainResized_i \subset \pixelDomainResized.
\)

\subsection{Factor Graph Optimization}
\label{sec:factor_graph_optimization}
At the core of our approach is the joint optimization over the dense depth variables
\(
\depth \in \Rgz^{\heightResized \times \widthResized}
\)
and the patch-wise alignment variables
\(
\slope, \bias \in \R^{\numPatches}
\). We denote \( \depth[p] \) as the depth at pixel position \( p \in \pixelDomainResized \) and \( \slope[i] \) and \( \bias[i] \) as the local affine alignment parameters of the \( i \)-th patch, representing the slope and the bias of the affine model.

\subsubsection{Cost Function}
\label{sec:factors-cost}
\begin{figure}
    \centering
    \includesvg[width=0.75\linewidth]{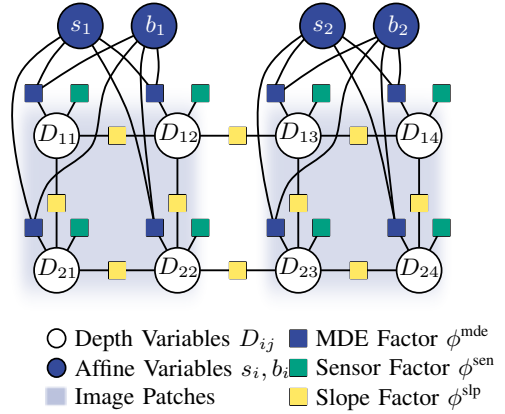}
    \caption{\textbf{Factor graph formulation. } %
    Dense per-pixel depth variables \(D_{ij}\) and patch-wise affine parameters \((s_i,b_i)\) are jointly optimized. Ternary MDE factors \(\phi^{\mathrm{mde}}\) align depth to the monocular prediction within each patch, unary sensor factors \(\phi^{\mathrm{sen}}\) enforce metric consistency, and binary logarithmic slope factors \(\phi^{\mathrm{slp}}\) preserve relative depth structure across pixels and patch boundaries. For simplicity, only four pixels per patch are shown.}
    \label{fig:factor-graph-overview}
    \vspace{-0.3cm}
\end{figure}

As depicted in \cref{fig:factor-graph-overview}, our factor graph consists of three types of factors, each encoding a complementary constraint on the optimization variables. 
Monocular depth alignment is enforced by a set of ternary factors \( \mdeFactor \) that couple per-pixel depth variables to patch-wise affine alignment parameters. 
Metric consistency is imposed by unary sensor factors \( \senFactor \), which anchor depth estimates to available sensor measurements. 
Finally, binary logarithmic slope consistency factors \( \slopeFactor \) regularize relative depth variations between neighboring pixels to follow those of the monocular depth estimation prior, preserving surface structure and object boundaries in regions with sparse or missing sensor data, including across patch boundaries.

The global objective is obtained by summing all factor contributions and weighting each factor type by a positive scalar \( \weightSymb^{\langle \cdot \rangle} \in \Rgz\). Formally, we minimize the total cost
\begin{align}
\cost(\depth, \slope, \bias) = \,
&\weightMde \sum_{i=1}^{\numPatches} \sum_{p \in \pixelDomainResized_i}
\mdeFactor(\depth[p], \slope[i], \bias[i]) \notag \\
&+ \weightSen \sum_{p \in \pixelDomainResized}
\senFactor(\depth[p]) \notag \\
&+ \weightSlope \sum_{(p, q) \in \neighbourSet}
\slopeFactor(\depth[p], \depth[q]).
\label{eq:total_energy}
\end{align}
where \( \neighbourSet \) denotes the set of 4-connected neighboring pixel pairs in \( \pixelDomainResized \).

\subsubsection{Monocular Depth Alignment Term}
\label{sec:mde_term}
The monocular depth estimation model provides a dense prior on scene geometry but predicts depth only up to an unknown affine transformation with respect to metric scale. To integrate this information into our factor graph, we introduce patch-wise monocular alignment factors \(\mdeFactor \) that relate each optimized depth variable \( \depth[p]\) to the corresponding monocular prediction \( \depthRescaledMde[p] \) through the local affine model \(\slope[i], \bias[i] \). For a given \( p \in \pixelDomainResized_i \) we define the factor as
\[
\mdeFactor(\depth[p], \slope[i], \bias[i]) =
\huber_{\huberThresholdMdeSen}\big(\depth[p] - (\slope[i] \depthRescaledMde[p] + \bias[i])\big),
\label{eq:mde-factor}
\]
where \( \huber \) is the robust huber cost~\cite{huber} with threshold \( \huberThresholdMdeSen \).
\subsubsection{Sensor Consistency Factor}
\label{sec:sensor_term}
The ternary monocular depth alignment factor alone yields infinitely many optimal solutions for a single pixel \( p \in \pixelDomainResized \). To ground the monocular prediction to metric space, we apply a huber cost \( \huber \) with the same threshold \( \huberThresholdMdeSen \) as in \cref{eq:mde-factor} between each optimized depth variable \( \depth[p] \) and the metric sensor measurement \( \depthRescaledSen[p] \). For missing depth values in the sensor depth map, we set the corresponding factor cost to zero:
\[
\senFactor(\depth[p]) = \huber_{\huberThresholdMdeSen}(\indicator{\depthRescaledSen[p] \ne \depthnan} \cdot (\depth[p] - \depthRescaledSen[p]))
\label{eq:sen-factor}
\]

\begin{figure}[t!]
    \centering
    \includesvg[width=\linewidth]{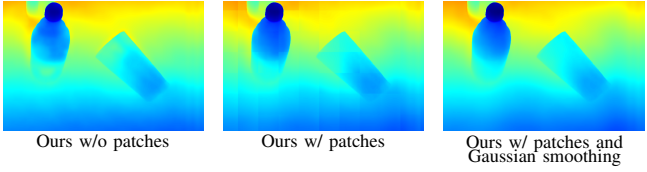}
    \caption{\textbf{Qualitative comparison of ablations.} We compare the outputs of the ablated variants (w/ and w/o patches) to \ourName{} where we additionally apply Gaussian smoothing on the patch-wise affine alignment parameters.}
    \label{fig:viz-patch}
    \vspace{-0.3cm}
\end{figure}

\subsubsection{Logarithmic Slope Consistency Factor}
\label{sec:slope_term}
Absolute depth alignment alone may fail in extremely sparse or noisy regions that are typically present in scenes containing non-Lambertian surfaces. We therefore additionally constrain relative depth changes in the optimized depth variables to follow those of the monocular estimate. For pixels \((p, q) \in \mathcal N\), where \( \mathcal N \) represents the set of 4-connected neighborhood within \( \pixelDomainResized \), we apply a robust cost \( \huber \) with threshold \( \huberThresholdSlope \) between the logarithmic slopes from the optimized depth map and the monocular depth estimation prior:
\begin{align}
\slopeFactor(\depth[p], \depth[q]) =
&\huber_{\huberThresholdSlope}\big( [\log \depth[p] - \log \depth[q]]\,-\notag \\
&[\log \depthRescaledMde[p] - \log \depthRescaledMde[q]] \big)
\label{eq:log-slope-consistency-factor}
\end{align}
Note, that  the log-space formulation in \cref{eq:log-slope-consistency-factor} yields scale-invariant slope constraints, since \(\log x y_1 - \log x y_2 = \log y_1 - \log y_2\) holds for any \( x, y_1, y_2 \in \Rgz \). 

\subsection{Optimization}
\label{sec:optimization}
We initialize the affine patch variables \( \slope \) and \( \bias \) via a global least-squares fit between valid sensor depth values and monocular predictions using \( \numSamplesInit \) randomly sampled pixels. The resulting scaled monocular depth serves as the initialization for \( \depth \). Starting from this initialization, we minimize the cost function in \cref{eq:total_energy} using Iterative Reweighted Least Squares (IRLS)~\cite{irls}, resulting in the optimized variables \( \depthOptimized \), \( \slopeOptimized \), and \( \biasOptimized \).

\subsection{Gaussian-Weighted Smoothing}
\label{subsec:gauss-weighted-smoothing}
As illustrated in \cref{fig:viz-patch}, directly using \(\depthOptimized\) as the final prediction can be disadvantageous for downstream tasks, since discontinuities remain at patch boundaries. To obtain a smooth final prediction, we compute pixel-wise affine parameters \(\slope[p]\) and \(\bias[p]\) for each \(p \in \pixelDomainResized\) by Gaussian-weighted averaging of the optimized patch-wise parameters \(\slopeOptimized[i]\) and \(\biasOptimized[i]\).
Let \(\patchCenter{i}\) denote the center of the \(i\)-th patch. We define the weight \(w_{p,i}\) based on the Euclidean distance between \(p\) and \(\patchCenter{i}\) using a Gaussian kernel \(g(\cdot)\) with a standard deviation of \(\sigma = m\) and normalize the weights such that \(\sum_{i=1}^n w_{p,i} = 1\):
\[
w_{p,i} =
\frac{\gauss{\|p-\patchCenter{i}\|_2^2}}
{\sum_{j=1}^{\numPatches} \gauss{\|p-\patchCenter{j}\|_2^2}}.
\]
The pixel-wise affine parameters are then given by
\[
\slope[p] = \sum_{i=1}^{\numPatches} w_{p,i}\,\slopeOptimized[i],
\qquad
\bias[p] = \sum_{i=1}^{\numPatches} w_{p,i}\,\biasOptimized[i].
\]
Applying the pixel-wise affine parameters to the rescaled monocular depth map \(\depthRescaledMde\) yields the predicted depth at pixel \(p\) in the resized image domain:
\[
\depthRescaledPredicted[p] = \slope[p]\,\depthRescaledMde[p] + \bias[p].
\]
Finally, we resize \(\depthRescaledPredicted \in \Rgz^{\heightResized \times \widthResized}\) back to the original resolution using nearest-neighbor interpolation, resulting in the final metric depth prediction \(\depthPredicted \in \Rgz^{\height \times \width}\). 
In practice, we implement the Gaussian-weighted averaging efficiently via Gaussian convolution on sparse parameter maps. Specifically, we construct sparse maps containing \(\slopeOptimized[i]\) and \(\biasOptimized[i]\) at the patch centers \(\patchCenter{i}\) and \(0\) elsewhere, and normalize the result using the convolution of the same Gaussian kernel with a sparse indicator map that is \(1\) at patch centers and \(0\) otherwise.

\section{Experimental Evaluation}
\label{sec:experiments}

In this section, we present both quantitative and qualitative results of \ourName{}. In \cref{subsec:quantitaive-evaluation}, we compare a set of methods on our introduced dataset \datasetName{} as well as the ClearPose dataset~\cite{clearpose}. We further outline the downstream applicability of \ourName{} on ScanNet++~\cite{scannetpp} in \cref{subsec:qualitative-results}.

\begin{table*}
    \centering
    \scriptsize
    \setlength\tabcolsep{8.5pt}
    \begin{tabular}{l|ccc|ccc|ccc}
        \toprule
        \multirow{3}{*}{Method} & \multicolumn{3}{c|}{Objects} & \multicolumn{3}{c|}{Background} & \multicolumn{3}{c}{Full} \\
        \cmidrule(lr){2-4} \cmidrule(lr){5-7} \cmidrule(lr){8-10}
         & 
            \makecell{MAE} & 
            \makecell{RMSE} & 
            \makecell{REL} & 
            \makecell{MAE} & 
            \makecell{RMSE} & 
            \makecell{REL} & 
            \makecell{MAE}  & 
            \makecell{RMSE} & 
            \makecell{REL} \\
        \midrule
        \ourNameTable[no-smooth] 
            & 0.025 & 0.045 & 0.042
            & 0.009 & 0.021 & 0.013
            & 0.011 & 0.024 & 0.016 \\
        \ourNameTable[no-patch-no-smooth] 
            & 0.029 & 0.050 & 0.049
            & 0.010 & 0.023 & 0.014
            & 0.012 & 0.027 & 0.018\\
        \midrule
        Inpainting 
            & 0.050 & 0.097 & 0.087 
            & \underline{0.011} & 0.092 & \underline{0.015} 
            & \underline{0.015} & 0.093 & \underline{0.023} \\
        Affine Alignment, e.g.~\cite{patel2025robotic, li2025novaflow}
            & \underline{0.034} & \underline{0.052} & \underline{0.059} 
            & 0.020 & \underline{0.031} & 0.028 
            & 0.022 & \underline{0.034} & 0.031 \\
        DepthAnything3-metric \cite{dav3} 
            & 0.088 & 0.106 & 0.153 
            & 0.079 & 0.105 & 0.102 
            & 0.080 & 0.106 & 0.107 \\
        \midrule
        \ourNameTable{} 
            & \textbf{0.026} & \textbf{0.045} & \textbf{0.043}
            & \textbf{0.010} & \textbf{0.021} & \textbf{0.013}
            & \textbf{0.011} & \textbf{0.025} & \textbf{0.016} \\
        \bottomrule
    \end{tabular}
    \caption{\textbf{Evaluation on \datasetName.} We report the Mean Absolute Error (MAE) ($\downarrow$) in meters, the Root Mean Squared Error (RMSE) ($\downarrow$) and the relative error (REL) ($\downarrow$). We distinguish between the depth values of objects and the background and report the full errors over all pixels. Best results are in \textbf{bold}, second best are \underline{underlined}. We observe that DAV3-metric~\cite{dav3} is already outperformed by affine alignment and even further by \ourName{}.}
    \label{tab:snp-results}
\end{table*}

\begin{table*}
    \centering
    \resizebox{\linewidth}{!}{
    \begin{tabular}{l|ccc|ccc|ccc|ccc|ccc}
        \toprule
        \multirow{3}{*}{Method} & \multicolumn{3}{c|}{New Background}
        & \multicolumn{3}{c|}{Heavy Occlusions}
        & \multicolumn{3}{c|}{Liquid}
        & \multicolumn{3}{c|}{Non-Planar}
        & \multicolumn{3}{c}{Opaque Objects} \\
        \cmidrule(lr){2-4} \cmidrule(lr){5-7} \cmidrule(lr){8-10} \cmidrule(lr){11-13} \cmidrule(lr){14-16}
         
            & \makecell{MAE} & \makecell{RMSE} & \makecell{REL}
            & \makecell{MAE} & \makecell{RMSE} & \makecell{REL}
            & \makecell{MAE} & \makecell{RMSE} & \makecell{REL}
            & \makecell{MAE} & \makecell{RMSE} & \makecell{REL}
            & \makecell{MAE} & \makecell{RMSE} & \makecell{REL} \\
        \midrule
        \ourNameTable[no-patch]
            & 0.079 & 0.247 & 0.090
            & 0.107 & 0.245 & 0.125
            & 0.085 & 0.272 & 0.102
            & 0.079 & 0.208 & 0.093
            & 0.083 & 0.272 & 0.104 \\
        \ourNameTable[no-smooth]
            & 0.078 & 0.232 & 0.090
            & 0.112 & 0.240 & 0.132
            & 0.081 & 0.251 & 0.098
            & 0.076 & 0.194 & 0.091
            & 0.079 & 0.243 & 0.100 \\
        \midrule
        Inpainting 
            & 0.165 & 0.417 & 0.191
            & 0.185 & 0.465 & 0.220
            & 0.167 & 0.397 & 0.206
            & 0.153 & 0.273 & 0.183
            & 0.144 & 0.346 & 0.186 \\
        Affine Alignment, e.g.~\cite{patel2025robotic, li2025novaflow}
            & \underline{0.105} & \underline{0.236} & \underline{0.123}
            & 0.138 & 0.251 & 0.165
            & \underline{0.099} & \underline{0.255} & \underline{0.122}
            & \underline{0.094} & \underline{0.196} & \underline{0.115}
            & \underline{0.096} & \textbf{0.244} & \underline{0.123} \\
        DepthAnything3-metric \cite{dav3}
            & 0.124 & \textbf{0.231} & 0.144
            & \textbf{0.109} & \textbf{0.216} & \textbf{0.131}
            & 0.103 & \textbf{0.240} & 0.126
            & 0.125 & 0.238 & 0.148
            & 0.128 & 0.278 & 0.166 \\
        \midrule
        \ourNameTable{}
            & \textbf{0.080} & \underline{0.234} & \textbf{0.091}
            & \underline{0.112} & \underline{0.242} & \underline{0.132}
            & \textbf{0.083} & \underline{0.255} & \textbf{0.101}
            & \textbf{0.075} & \textbf{0.193} & \textbf{0.089}
            & \textbf{0.081} & \underline{0.246} & \textbf{0.102} \\
        \bottomrule
    \end{tabular}
    }
    \caption{\textbf{Evaluation on ClearPose\cite{clearpose}.} We compare \ourName{} across the dataset splits of ClearPose~\cite{clearpose} and employ the same metrics as in \cref{tab:snp-results}.}
    \label{tab:cp-results}
\end{table*}
\begin{figure*}
    \centering
    \includesvg[width=\textwidth]{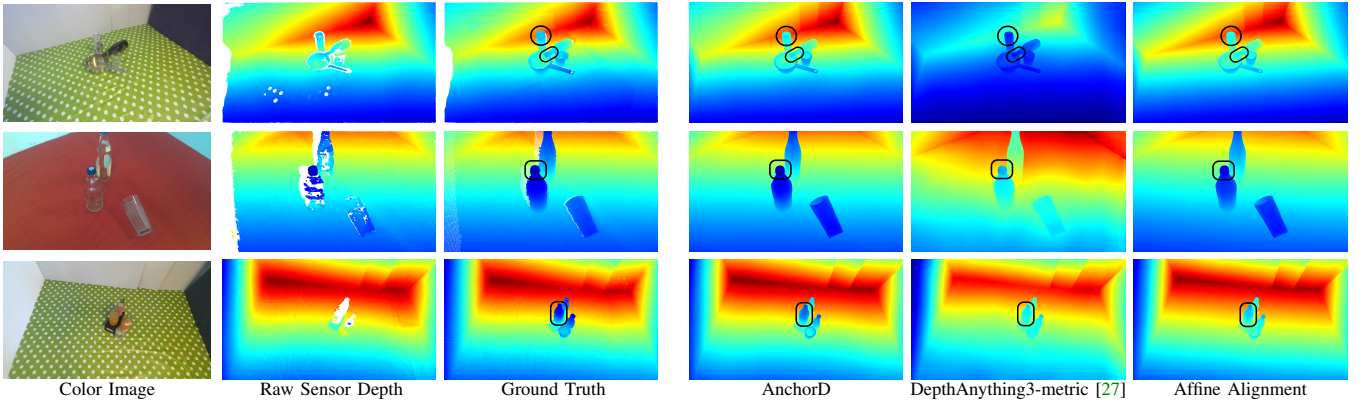}
    \caption{\textbf{Qualitative results on \datasetName{}.} We compare \ourName{} (ours) with metric Depth Anything3~\cite{dav3} as well as affine scaling in the three rightmost columns. For reference, we display the raw sensor depth and ground truth depth in the two leftmost columns. All depth images along each row feature a consistent color scaling.}
    \label{fig:qualitative2D-snp}
    \vspace{-0.3cm}
\end{figure*}

\subsection{Quantitative Evaluation}
\label{subsec:quantitaive-evaluation}

\noindent\textbf{Datasets:} We evaluate on \datasetName{} and ClearPose~\cite{clearpose}. On \datasetName{}, we report results at a resolution of \(720 \times 1280\) on object regions, complementary background regions, and the full image.
For ClearPose, we evaluate at a resolution of \(480 \times 640\) and restrict the evaluation to object masks, as the dataset does not provide dense ground truth annotations for the full scene. We report results on the fine-grained test categories \textit{New Background}, \textit{Heavy Occlusion}, \textit{Opaque Distractor}, \textit{Filled Liquid}, and \textit{Non Planar}. We exclude the \textit{Translucent Cover} category as its annotation protocol conflicts with our workspace-depth definition in regions where objects are visible through translucent covers (see \cref{sec:dataset}). Since the originally used evaluation splits are not publicly available, we sample 2000 random frames for each category. We use the same hyperparameter setting as for \datasetName{}, except that we reduce the patch size to \(m = 48\), which preserves a consistent number of patches along the vertical image dimension.

\noindent{\textbf{Implementation Details:}} 
Our approach employs jaxLS, enabling efficient parallel execution on the GPU. We utilize a pre-trained monocular DepthAnything3 model for MDE predictions~\cite{dav3}, a patch size of $\patchSize=64$ on \datasetName{} and $\patchSize=48$ on ClearPose~\cite{clearpose}, respectively. We choose $\weightMde=2.5$, $\weightSen=0.5$, $\weightSlope=1.0$, $\delta_1 = 0.002$, $\delta_2 = 0.01$, and \(\numSamplesInit = 64\).\looseness=-1

\noindent\textbf{Metrics:}
As standard for depth completion, we evaluate quantitative performance using mean absolute error (MAE), root mean squared error (RMSE), and mean relative error (REL).\looseness=-1

\noindent\textbf{Baselines:}
We compare \ourName{} against three baselines. First, we consider a classical inpainting method based on the Navier--Stokes formulation of image inpainting~\cite{inpainting}, which fills missing values in the raw sensor depth map \(\depthSen\) without using RGB information. This baseline isolates the contribution of purely depth-based hole filling. Second, we evaluate a global affine least-squares baseline that aligns the monocular depth prediction \(\depthMde\) to the sensor depth \(\depthSen\) using 64 randomly sampled valid pixels, as already commonly used~\cite{patel2025robotic}. This baseline assesses whether a single global affine transformation is sufficient to map the monocular prediction into metric space~\cite{li2025novaflow}. Currently, this approach represents the de-facto standard in utilizing metric/monocular depth foundation models in robotic deployments. Third, we compare with the pretrained DepthAnything3-metric model~\cite{dav3}. Following the recommendation of the original work, we scale its output by \(f_x / 300\), where \(f_x\) denotes the camera focal length in pixels.

\begin{figure*}
    \centering
    \includesvg[width=\textwidth]{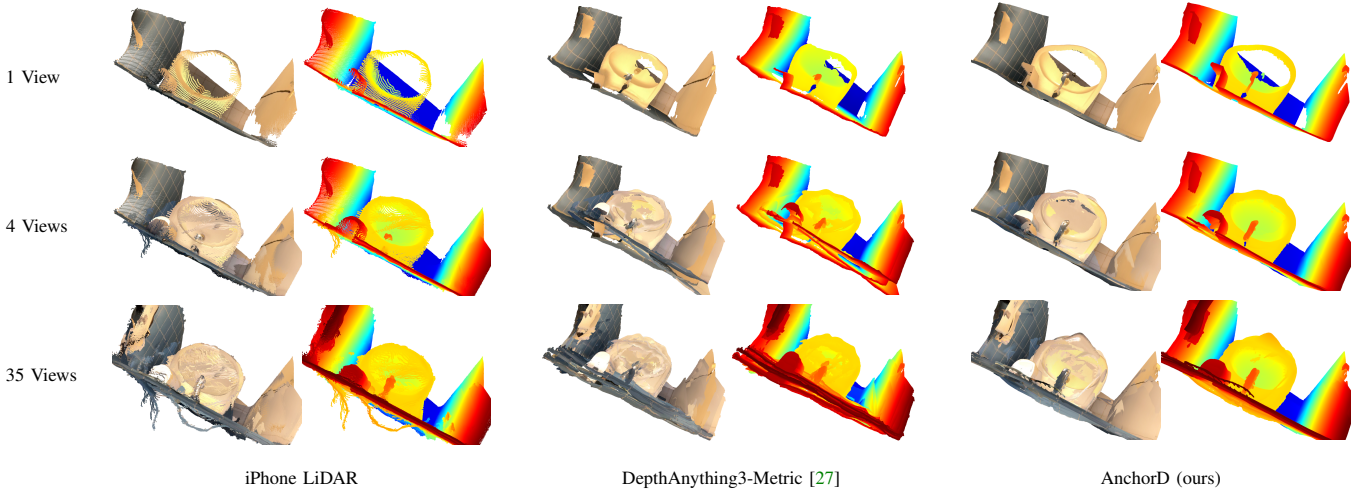}
    \caption{\textbf{Qualitative results on ScanNet++.} We compare the point cloud aggregation quality when aggregating the depth over multiple camera poses stemming from the raw iPhone LiDAR observations (left), a metric foundation model (DepthAnything3) taking RGB as input (middle), and \ourName{} (ours) (right).}
    \label{fig:scannet-3d-viz}
\end{figure*}

\noindent\textbf{Ablations:}
To analyze the contribution of the individual components of \ourName{}, we consider two ablated variants. In \ourName{}-no-smooth, we remove the Gaussian smoothing step and directly use the optimized depth \(\depthOptimized\) as the final prediction. In \ourName{}-no-patch, we remove the patch-wise decomposition and instead use a single global patch in the factor graph. As in the previous ablation, the final prediction is given directly by \(\depthOptimized\).

\noindent\textbf{Results:}
We present results on \ourName{} in \cref{tab:snp-results} and on ClearPose~\cite{clearpose} in \cref{tab:cp-results}. On \datasetName{}, \ourName{} outperforms all baseline methods across all metrics and evaluation regions. The improvement is especially evident on object regions, confirming that the proposed optimization framework effectively corrects corrupted depth where monocular priors or simple affine alignment alone are insufficient. The ablation study shows that the patch-wise parameterization is beneficial, as removing it consistently degrades performance, whereas the smoothing step has only a minor influence on the reported error metrics, as intended, and mainly acts as a regularizer for spatial consistency (see \cref{fig:viz-patch}).

On ClearPose, \ourName{} confirms the same overall trend and achieves the strongest performance across the dataset splits, obtaining the best MAE and REL in four out of five categories. The only exception is \textit{Heavy Occlusions}, where the DepthAnything3-metric baseline performs marginally better. These findings indicate that the local patch-wise alignment effectively transfers to diverse transparent-object scenarios and reduces the typical depth error more reliably than global affine alignment or metric depth foundation models. While RMSE gains are less consistent, \ourName{} remains competitive in all categories, suggesting that the remaining errors are dominated by a small number of severe failure cases that disproportionately influence this metric.

\begin{figure}
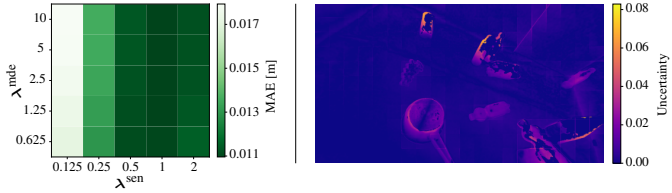

\centering
\footnotesize
\begin{minipage}[t]{0.40\linewidth}
  \vspace{0pt}\centering
  \includesvg[width=\linewidth]{assets/figures/sensitivity.svg}
\end{minipage}
\hfill
\rule[-2.11cm]{0.3pt}{2.11cm}
\hfill
\begin{minipage}[t]{0.54\linewidth}
  \vspace{0pt}\centering
  \includesvg[width=\linewidth]{assets/figures/residuals.svg}
\end{minipage}
\caption{\textbf{Model Sensitivity and Uncertainty. } We display a parameter sensitivity matrix delineating the influence of various combinations of \(\weightMde\) and \(\weightSen\) on the resulting global MAE (left) and a normalized residual map representing our model's uncertainty across image regions for the sample from \cref{fig:approach-overview} (right).} 
\label{fig:sensitivity_and_residual}
\end{figure}

\noindent\textbf{Hyperparameter Sensitivity:} We vary \(\weightMde\) and \(\weightSen\) around the default on the full image region to assess sensitivity (see \cref{fig:sensitivity_and_residual}). We find that performance is mainly driven by the unary sensor weight \(\weightSen\), while results are robust to \(\weightMde\) perturbations. Although \(\weightSen \approx 1\) performs slightly better on the full image region, we opt for \(0.5\) to downweight erroneous sensor measurements in transparent regions.

\subsection{Qualitative Results}
\label{subsec:qualitative-results}

\noindent\textbf{2D on \datasetName{}:}
We show qualitative results of the predicted depth maps in \cref{fig:qualitative2D-snp}.
All methods preserve similar depth discontinuities since they are built on the same geometric monocular prior. Thus, the overall coarse structure is similar, but the recovery of absolute metric depth differs. In particular, DepthAnything3-metric~\cite{dav3} often mis-scales the scene globally, likely in cases where RGB cues alone provide insufficient information about overall scale. In contrast, \ourName{} appears especially beneficial in small and transparent object regions, where the monocular prior is often locally distorted. This supports our claim that the patch-wise factor-graph formulation can locally correct erroneous monocular predictions while preserving the global geometric structure.

\noindent\textbf{3D on ScanNet++:}
We also evaluate the overall consistency of our predictions when aggregating depth maps into point clouds in 3D on ScanNet++~\cite{scannetpp}. To do so, we compare the output of \ourName{} against the low-resolution iPhone depth, and depth maps from the metric variant of DepthAnything3~\cite{dav3}. 
Visualizations (see \cref{fig:scannet-3d-viz}) show that \ourName{} yields the most favorable trade-off between density and geometric consistency. Compared to the sparse iPhone LiDAR depth, our method yields substantially denser reconstructions while preserving a similarly coherent global scene geometry. In particular, \ourName{} is able to recover small non-Lambertian structures that are missing in the raw sensor measurements, such as the sink surface in the single-frame reconstruction. In comparison, the metric variant of DepthAnything3~\cite{dav3} exhibits noticeably weaker cross-view consistency: even with a small number of viewpoints, large planar surfaces, such as walls, are not reliably aligned, leading to visible geometric inconsistencies in the aggregated 3D reconstruction.\looseness=-1

\noindent\textbf{Uncertainty Estimate:} We illustrate uncertainty using the pixel-wise weighted sum of the unary sensor residual \(\senFactor\) and the ternary MDE residual \(\mdeFactor\), normalized by the maximum pixel-wise residual over the full dataset in \cref{fig:sensitivity_and_residual}. We find that background regions are generally well aligned, while uncertainty is concentrated in transparent objects, mainly due to ambiguous and noisy sensor measurements.

\section{Conclusion}
\label{sec:conclusion}
We presented \ourName{}, a training-free depth completion framework that fuses raw sensor depth with monocular depth priors via factor graph optimization, recovering dense metric depth on transparent and non-Lambertian surfaces without task-specific training. Patch-wise affine alignment locally grounds monocular predictions in metric space, enabling generalization across sensors and domains. Additionally, we introduced the real-world \datasetName{} benchmark with dense scene-wide ground truth depth for non-Lambertian scenes, addressing a key gap in existing evaluation protocols. Future work will examine the downstream impact on grasping and manipulation tasks.

\begin{footnotesize}
    \bibliographystyle{IEEEtran}
    \bibliography{sources.bib}
\end{footnotesize}

\end{document}